\documentclass{article}
\usepackage[utf8]{inputenc}
\usepackage{a4wide}
\usepackage{xcolor}
\usepackage{amsmath}
\usepackage{lscape}
\usepackage[font=small,labelfont=bf]{caption}
\usepackage{graphicx}
\usepackage{subcaption}
\usepackage{comment}
\usepackage{tabularx}
\usepackage{multirow}
\usepackage{algorithmic}
\usepackage[ruled,vlined]{algorithm2e}
\usepackage{amsmath,amssymb,amsfonts,amsthm} 
{

\theoremstyle{remark}

}
\usepackage[utf8]{inputenc}
\usepackage{amssymb}
\usepackage{multicol}
\usepackage{geometry}
\usepackage{hyperref}
\usepackage{authblk}
\usepackage{ulem}
\usepackage{moreverb}
\usepackage{amsmath,bm}
\usepackage{graphicx}
\usepackage{siunitx}
\usepackage{subcaption}

\renewenvironment{abstract}
  {\small\noindent\textbf{Abstract.}\normalsize\ignorespaces}
  {\par\noindent\ignorespacesafterend}



\begin{document}

\title{Position-Based Flocking for Persistent Alignment without Velocity Sensing}

\author[1]{Hossein B. Jond}
\author[2]{Veli Bakırcıoğlu}
\author[3]{Logan E. Beaver}
\author[4,5,6]{Nejat Tükenmez}
\author[7]{Adel Akbarimajd}
\author[1]{Martin Saska}
\affil[1]{Department of Cybernetics, Czech Technical University in Prague, Prague, Czechia}
\affil[2]{Aksaray University, 68100, Aksaray, Türkiye}
\affil[3]{Mechanical and Aerospace Engineering, Old Dominion University, Norfolk, VA, USA}
\affil[4]{Mechatronics Engineering Department, Istanbul Technical University, Istanbul, Turkiye}
\affil[5]{The Daniel Guggenheim School of Aerospace Engineering, Georgia Institute of Technology, Atlanta, GA, USA}
\affil[6]{Mechatronics Engineering Department, Isparta University of Applied Sciences, Isparta, Turkiye}
\affil[7]{Department of Electrical and Computer Engineering, University of Mohaghegh Ardabili, Ardabil, Iran}

\date{}

\maketitle

\begin{abstract}
Coordinated collective motion in bird flocks and fish schools inspires algorithms for cohesive swarm robotics. This paper presents a position-based flocking model that achieves persistent velocity alignment without velocity sensing. By approximating relative velocity differences from changes between current and initial relative positions and incorporating a time- and density-dependent alignment gain with a non-zero minimum threshold to maintain persistent alignment, the model sustains coherent collective motion over extended periods. Simulations with a collective of 50 agents demonstrate that the position-based flocking model attains faster and more sustained directional alignment and results in more compact formations than a velocity-alignment-based baseline. This position-based flocking model is particularly well-suited for real-world robotic swarms, where velocity measurements are unreliable, noisy, or unavailable. Experimental results using a team of nine real wheeled mobile robots are also presented.
\end{abstract}

{\textbf{Keywords:}{ Position-based flocking, Persistent alignment, Collective motion, Swarm robotics, Velocity-free alignment, Bio-inspired coordination}

\section{Introduction}\label{intro}

Collective behaviors, such as flocking behavior, observed in natural systems such as bird flocks, fish schools, and insect swarms, represent a fascinating example of emergent collective dynamics \cite{reynolds1987flocks}. These systems exhibit coordinated motion, driven by local interactions among individuals. Understanding and replicating such bio-inspired behaviors in artificial systems has significant implications for robotics, autonomous vehicle coordination, and swarm intelligence \cite{viragh2014flocking}.

Coordinated movements in bird flocks and fish schools emerge from local social interactions, yet their developmental origins, neurobiological basis, and the emergence of group-level dynamics remain a core challenge in collective behavior research. Traditional models assume animals follow simple behavioral rules. Reynolds' Boids \cite{reynolds1987flocks}, introduced three core behavioral rules: cohesion (staying close to neighbors), separation (avoiding collisions), and alignment (matching velocities). These principles have inspired numerous mathematical and computational models to capture the dynamics of collective motion \cite{vicsek1995novel,cucker2007emergent,olfati2006flocking}. The Vicsek model, for instance, focuses on alignment through local velocity averaging, demonstrating phase transitions from disordered to ordered motion. However, such models often rely on velocity information, which is difficult to measure directly in robotic systems or requires error-prone estimation due to sensor noise \cite{schilling2021vision,wang2023collective}. Thus, collective motion models based solely on position data are both practical and significant for robust coordination. 

Recent empirical studies challenge the necessity of explicit velocity alignment in collective biological motion. \cite{murakami2017emergence} demonstrate that swarms achieve high polarity through mutual anticipation, where individuals adjust directions based on anticipated neighbor moves without direct velocity matching. \cite{xiao2024perception} show that birds regulate heading and acceleration through perceived motion salience of neighbors, rather than explicit alignment. \cite{salahshour2025allocentric} proposes that collective motion spontaneously emerges from neurobiological navigational circuits when individuals use allocentric bearings to neighbors, without requiring dedicated alignment rules. Similarly, \cite{zada2024development} find that micro glassfish develop visual postural alignment and positional aggregation solely through recognition of conspecific shape and posture. These bio-inspired mechanisms highlight that coherent motion arises from alternative interactions beyond traditional alignment rules, such as persistent spatial relationships. 
These findings collectively suggest that spatial relationships and anticipatory positional cues may be more fundamental than explicit velocity matching in biological collectives.

Velocity estimation is biologically implausible, whereas relative displacement over time is naturally available through vision. Position-based flocking models circumvent the need for explicit velocity measurements by using relative positions to define interaction rules \cite{zhan2012flocking}. These models are particularly well-suited to systems with readily available position data (e.g., GPS-equipped drones or sensor networks) \cite{viragh2014flocking}, and are widely adopted in robotic multi-agent systems due to their simplicity, robustness, computational efficiency, scalability, and fault tolerance.

Alignment—essential for achieving velocity consensus and collective motion coherence—can also be effectively achieved from position. By approximating velocity differences from sampled or relative position data, these rules maintain robustness in noisy environments and significantly reduce communication requirements. For second-order (inertial) agents, position-based protocols have been proven capable of achieving consensus even without direct velocity sensing or communication \cite{mei2013distributed,yu2011second}, and they remain effective under time-varying topologies as long as the network remains strongly connected \cite{gao2009consensus}.

This paper builds on the velocity-alignment-based flocking scheme of \cite{jond2025minimal} and elaborates on the position-based flocking model introduced in \cite{jond2025position}. By approximating relative velocity differences using initial and current positions and incorporating a non-vanishing threshold on the decaying alignment gain, the model achieves persistent emergent alignment and induces self-organized rigidity, leading to compact and stable formations. This weak memory effect—realized solely through the initial relative position history—provides a practical and biologically plausible alternative to instantaneous velocity sensing. To assess performance, we define a cosine-based alignment metric and track it alongside inter-agent separation, providing interpretable indicators of flocking quality. An ablation study (with vs. without the threshold) qualitatively demonstrates the critical role of the proposed threshold in sustaining coherent alignment. The resulting approach is particularly valuable for communication- and sensing-constrained swarm robotics that rely on relative position measurements (e.g., via UWB or LiDAR) and require bandwidth-efficient coordination. We demonstrate its effectiveness through simulations and experiments with real wheeled mobile robots.

An earlier version of this research was presented in \cite{jond2025position}. The present paper extends that version by (1) making the bio-inspiration underlying the position-based alignment more explicit, (2) providing a deeper biological interpretation of the results, (3) strengthening the relevance to robotics applications, and (4) including experimental validation using a team of real wheeled mobile robots. 

This paper is organized as follows. Section \ref{sec:baseline} presents the baseline velocity-alignment model. Section \ref{sec:position-based-flocking} introduces the position-based formulation. Section \ref{sec:numerical-simulations} provides simulation and comparative analysis. Section \ref{sec:experiment} reports experimental results with real robots. Section~\ref{sec:conclusion} concludes the paper.

\section{Flocking from Position and Velocity States}\label{sec:baseline}

The baseline velocity-alignment-based flocking model, adapted from~\cite{jond2025minimal}, governs agent interactions from position and velocity states. The model is described as follows.

Consider \(n \geq 2\) agents, indexed by \(\mathcal{N} = \{1, \dots, n\}\) in \(\mathbb{R}^d,d=\{2,3\}\). Each agent \(i \in \mathcal{N}\) has position \(\mathbf{p}_i\), velocity \(\mathbf{v}_i\), and acceleration/control \(\mathbf{u}_i\) in \(\mathbb{R}^d\). 

The agents move freely, with dynamics given by
\begin{align}\label{eq:flock-model}
    \dot{\mathbf{p}}_i = \mathbf{v}_i, \quad
    \dot{\mathbf{v}}_i = \mathbf{u}_i, 
\end{align}
for \(i \in \mathcal{N}\). The control input for agent \(i\in\mathcal{N}\) is
\begin{equation}\label{eq:control-flk}
    \mathbf{u}_i =\sum_{j \in \mathcal{N}_i} \psi(\|\mathbf{p}_j - \mathbf{p}_i\|)(\mathbf{p}_j - \mathbf{p}_i) +\sum_{j \in \mathcal{N}_i} (\mathbf{v}_j - \mathbf{v}_i),
\end{equation}
where the first term is an attractive-repulsive term to ensure cohesion-separation, and the second term enforces velocity alignment. The cohesion-separation gain is given by
\begin{align*}
    &\psi(\|\mathbf{p}_j - \mathbf{p}_i\|) = 1 - \frac{\delta_i|\mathcal{N}_i|}{\|\mathbf{p}_j - \mathbf{p}_i\|},\quad \|\mathbf{p}_j - \mathbf{p}_i\|>0, 
\end{align*}
where \(\delta_i\geq0\) is the (desired/equilibrium) distance scale and \(|\mathcal{N}_i|\) denotes the neighborhood size. 

The model achieves balanced cohesion–separation through the distance-dependent gain \(\psi\) and velocity consensus through the implicit gain (equal to one). Specifically, $\psi<0$ (repulsive) when $\|\mathbf{p}_j - \mathbf{p}_i\| < \delta_i |\mathcal{N}_i|$ ensures short-range repulsion; $\psi>0$ (attractive) when $\|\mathbf{p}_j - \mathbf{p}_i\| > \delta_i |\mathcal{N}_i|$ promotes long-range attraction/cohesion; and $\psi=0$ at $\|\mathbf{p}_j - \mathbf{p}_i\| = \delta_i |\mathcal{N}_i|$ marks equilibrium spacing. The distance scale $\delta_i$ directly tunes formation density: smaller $\delta_i$ yields compact flocks via shorter equilibrium spacing, while larger $\delta_i$ produces looser flocks. Short-range repulsion inherently prevents collisions for any finite \(\delta_i > 0\). The neighborhood size \(|\mathcal{N}_i|\) scales the equilibrium spacing, pushing neighbors outward as connectivity increases and preventing cohesion forces from collapsing densely connected agents. Together, \(r_i\) and \(|\mathcal{N}_i|\) jointly regulate the spatial structure of agent interactions, allowing the equilibrium configuration to adapt to the local topology of the interaction graph.

Velocity and control inputs are obtained via smooth saturation as
\begin{equation}  
\mathbf{x}_i = x_i^{\max} \tanh\left(\frac{\|\mathbf{x}_i^{\text{cmd}}\|}{x_i^{\max}}\right) \frac{\mathbf{x}_i^{\text{cmd}}}{\|\mathbf{x}_i^{\text{cmd}}\|},  
\end{equation}  
where \(\mathbf{x}_i^{\text{cmd}} \in \{\mathbf{v}_i^{\text{cmd}}, \mathbf{u}_i^{\text{cmd}}\}\) is the commanded (unsaturated) input, \(\mathbf{x}_i\in \{\mathbf{v}_i, \mathbf{u}_i\}\) is the applied (saturated) input and \(x_i^{\max} \in \{v_i^{\max}, u_i^{\max}\}\) is the corresponding maximum  magnitude (with the convention \(\mathbf{0}/\|\mathbf{0}\| = \mathbf{0}\)).

\section{Position-Based Flocking}\label{sec:position-based-flocking}

At \(t \to \infty\), \(\sum_{j \in \mathcal{N}_i} (\mathbf{v}_j - \mathbf{v}_i) \to 0\), implying velocity alignment and stable relative positions, since \(\frac{d}{dt} \sum_{j \in \mathcal{N}_i} (\mathbf{p}_j - \mathbf{p}_i) \to 0\).

Using the definition of the derivative, the velocity alignment term can be rewritten as
\[
\sum_{j \in \mathcal{N}_i} (\mathbf{v}_j - \mathbf{v}_i) = \frac{d}{dt} \left( \sum_{j \in \mathcal{N}_i} (\mathbf{p}_j - \mathbf{p}_i) \right).
\]

For a purely position-based rule, we approximate the relative velocity by the long-term change in relative position
\[
\mathbf{v}_j - \mathbf{v}_i \approx \frac{ (\mathbf{p}_j - \mathbf{p}_i) - (\mathbf{p}_j(0) - \mathbf{p}_i(0)) }{t},\quad t>0.
\]

Substituting yields the position-based approximation of the velocity alignment term
\begin{align}\label{eq:pos-ali}
    &\sum_{j \in \mathcal{N}_i}(\mathbf{v}_j - \mathbf{v}_i)\approx \frac{1}{t}\sum_{j \in \mathcal{N}_i} \left[ (\mathbf{p}_j - \mathbf{p}_i) - (\mathbf{p}_j(0) - \mathbf{p}_i(0)) \right].
\end{align}

Here, \eqref{eq:pos-ali} suggests that alignment may be promoted using only positional information, by reconstructing relative motion from the accumulated change in relative positions over time. The mechanism mirrors how biological swarms coordinate without direct access to metric velocities, inferring motion instead through local perceptual cues (see Fig.~\ref{fig:analogy}).

\begin{figure}
    \centering
    \includegraphics[width=0.5\linewidth]{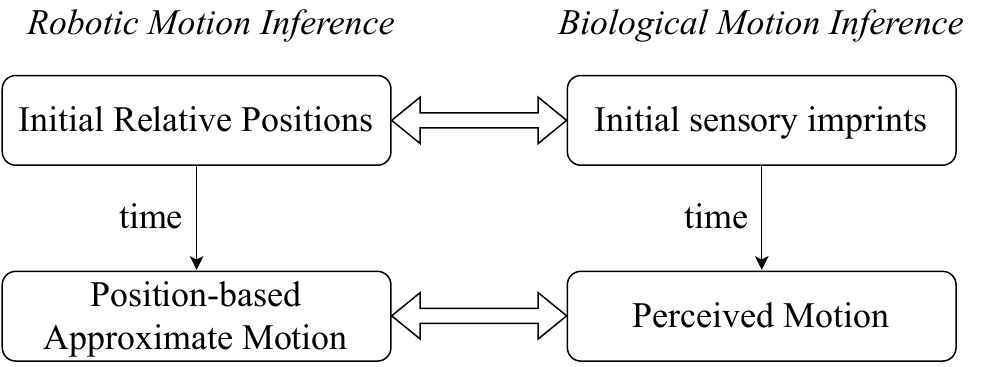}
    \caption{Conceptual analogy between motion inference in robotic and biological swarms. Robots reconstruct relative motion from time-integrated position differences, while biological agents infer motion directly from local sensory imprints.}
    \label{fig:analogy}
\end{figure}

Fig.~\ref{fig:conceptual} contrasts classical velocity alignment with the position-based mechanism. In the classical case, agents directly sense or exchange instantaneous relative velocities \(\mathbf{v}_j - \mathbf{v}_i\) to reach directional consensus. In the position-based approach, relative velocity is instead inferred from the temporal change in relative positions, using the initial configuration \(\mathbf{p}_j(0) - \mathbf{p}_i(0)\) as a weak \textit{imprinting}. This reference affects alignment strength mainly during early transients and gradually fades as \(t\) increases.

\begin{figure}
    \centering
    \begin{subfigure}{0.49\textwidth}
        \centering
        \includegraphics[width=0.7\linewidth]{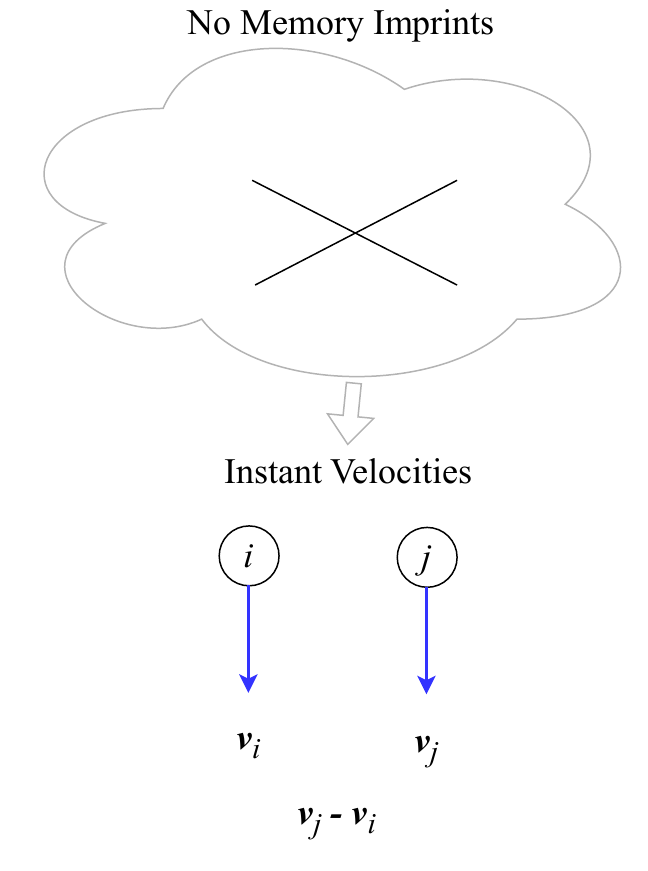}
        \caption{}
    \end{subfigure}
    \begin{subfigure}{0.49\textwidth}
        \centering
        \includegraphics[width=0.7\linewidth]{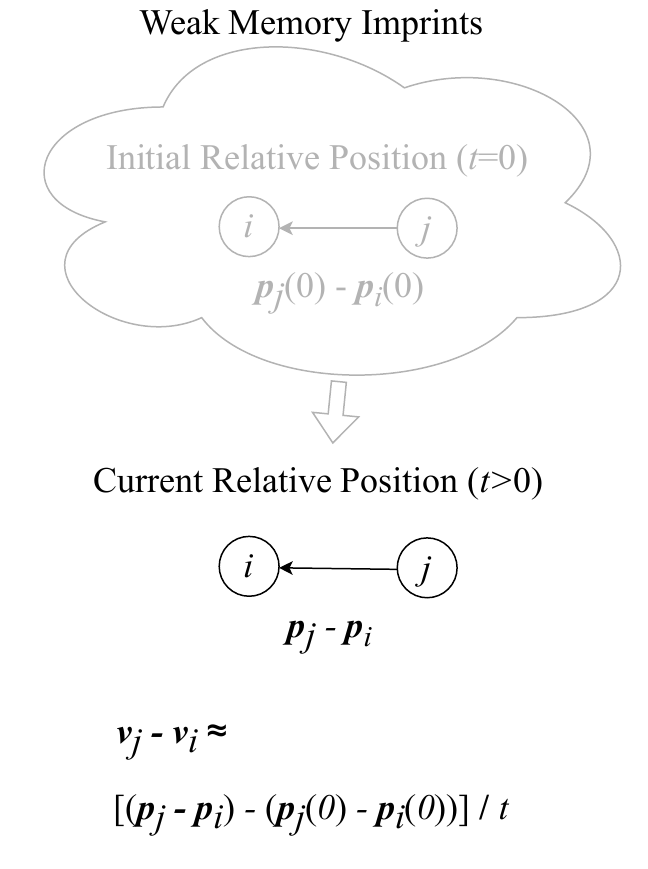}
        \caption{}
    \end{subfigure}
    \caption{(a) Velocity alignment: direct use of instantaneous \(\mathbf{v}_j - \mathbf{v}_i\) and (b) Position-based alignment: inference via \([(\mathbf{p}_j-\mathbf{p}_i)-(\mathbf{p}_j(0)-\mathbf{p}_i(0))]/t\).}
    \label{fig:conceptual}
\end{figure}

\subsection{Threshold-based persistence of alignment}

The gain $  \frac{1}{t}  $ in \eqref{eq:pos-ali} drives strong alignment early on but decays to zero as $  t \to \infty  $, preventing sustained coordination. At small $  t  $, limited accumulated displacement causes the term to underestimate relative velocity; as velocities converge, the fraction increasingly misrepresents the true alignment signal. We therefore introduce a threshold mechanism that replaces the decaying $  \frac{1}{t}  $ gain with a constant value once $  t > \frac{1}{k}  $, ensuring persistent velocity alignment while preserving biologically plausible transient dynamics.

The modified position-based alignment term takes the form
\begin{equation*}
\phi \sum_{j \in \mathcal{N}_i} \Bigl[ (\mathbf{p}_j - \mathbf{p}_i) - (\mathbf{p}_j(0) - \mathbf{p}_i(0)) \Bigr],
\end{equation*}
with the time-dependent gain
\[
\phi =
\begin{cases}
\dfrac{|\mathcal{N}_i|}{t}  & \text{if } 0< t \leq \dfrac{1}{k}, \\[1em]
k \, |\mathcal{N}_i|      & \text{if } t > \dfrac{1}{k},
\end{cases}
\]
where \(k > 0\) is a positive constant (e.g., \(k = 0.1\)). Furthermore, it moderates the initial amplification by \(|\mathcal{N}_i|\) in the transient phase.

As illustrated in Fig.~\ref{fig:phi}, this piecewise definition produces two distinct behavioral regimes. For small \(t\) (\(0<t \leq 1/k\)), the influence decays as \(|\mathcal{N}_i|/t\), yielding rapid, transient coordination reminiscent of short-term local adjustments in loosely coupled biological groups. Beyond the threshold $t = 1/k$, $\phi$ saturates at the constant steady-state gain $k|\mathcal{N}_i|$, enabling sustained alignment that is no longer sensitive to the accumulated relative displacements.

This qualitative transition from transient to persistent collective motion mirrors social reinforcement mechanisms observed in animal groups, where repeated local interactions progressively strengthen alignment tendencies and stabilize flock coherence once a critical cumulative interaction threshold is exceeded.

\begin{figure}
    \centering
    \includegraphics[width=0.5\linewidth]{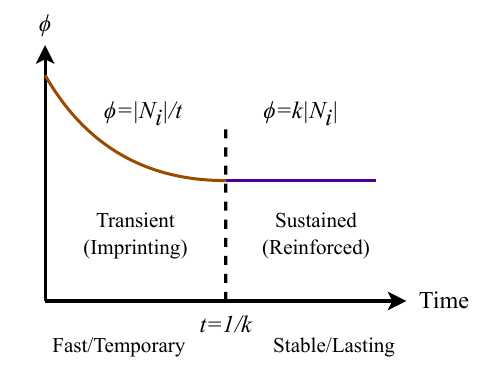}
    \caption{Time-dependent alignment gain \(\phi\). The function transitions from a decaying transient regime (\(\phi = |\mathcal{N}_i|/t\)) to a persistent regime (\(\phi = k|\mathcal{N}_i|\)) at the threshold \(t = 1/k\). Initial relative positions exert a strong influence (imprinting effect) before the threshold, after which alignment stabilizes (social reinforcement phase).}
    \label{fig:phi}
\end{figure}

The resulting position-based flocking input is
\begin{align}\label{eq:position-model}
    \mathbf{u}_i&=\sum_{j\in\mathcal{N}_i}\left(\psi(\lVert \mathbf{p}_j-\mathbf{p}_i\rVert)+\phi\right)(\mathbf{p}_j-\mathbf{p}_i)-\phi\sum_{j\in\mathcal{N}_i}(\mathbf{p}_j(0)-\mathbf{p}_i(0)),
\end{align}
where the influence of current and initial positions depends on the gain \(\phi\).

The key distinguishing features between velocity alignment and the position-based mechanism with memory imprinting are summarized in Table~\ref{tab:flocking-comparison}.
\begin{table}[t]
\caption{Velocity-based and position-based alignment components.}
\label{tab:flocking-comparison}
\centering
\begin{tabular}{|l|l|}
\hline
\textbf{Velocity-Based} & \textbf{Position-Based Alignment} \\
\hline
Direct relative velocity input & Inferred relative velocity via positions\\ 
 $\mathbf{v}_j-\mathbf{v}_i$ &  $[(\mathbf{p}_j-\mathbf{p}_i)-(\mathbf{p}_j(0)-\mathbf{p}_i(0))]/t$\\
\hline
 No memory of initial state &Weak memory imprint $\mathbf{p}_j(0)-\mathbf{p}_i(0)$\\
\hline
 Constant influence over time & Time-dependent behavioral switch \\
 &transient $(t\leq1/k)\rightarrow$ sustained $(t>1/k)$ \\
\hline
\end{tabular}
\end{table}

This position-based alignment term initially drives current relative positions toward the initial relative configuration of neighbors, inducing a short-lived shape memory effect. The collective briefly preserves the initial spacing pattern, resembling displacement-based formation behavior. However, this tendency interacts with the cohesion–separation mechanism, which attracts agents toward equilibrium distances, preventing sustained initial rigid formation.

Although this initial anchoring resembles classical displacement-based formation control \cite{olfati2002graph,cai2014adaptive}, the objective here is not to enforce a prescribed rigid formation. Instead, the initial configuration acts as a weak, transient memory imprint that gradually loses its influence, after which the alignment term becomes proportional to the current relative positions, promoting cohesion and a velocity-consensus-like flocking behavior. Thus, the mechanism is better interpreted as a bio-inspired, position-only approximation of Reynolds-style alignment rather than an instance of rigid formation control.

For robotic MAS without direct velocity sensing or communication, the proposed position-based flocking model \eqref{eq:position-model} substantially simplifies perception requirements (e.g., relying solely on LIDAR or range-only sensors), reduces computational load, and enhances robustness to noise. By inferring alignment from relative position changes over time, the approach exhibits coordination persistence even when velocity estimates are unavailable, noisy, or inconsistent. This enables fault-tolerant alignment under sensing or communication failures and scales favorably to larger teams, as only positional data is exchanged. In communication-enabled systems, the bandwidth demand is effectively reduced compared to position–velocity exchange, contributing to robustness, scalability, and a reduced infrastructure burden.

\section{Simulation} \label{sec:numerical-simulations}

To evaluate  the position-based flocking model \eqref{eq:position-model}, we simulate \(n = 50\) agents in a 2D space (\(d = 2\)) over \(t \in [0, 100]\, \text{s}\). Initial positions \(\mathbf{p}_i(0)\) are randomly distributed in a \(25 \times 25\, \text{m}\) square, with initial velocities \(\mathbf{v}_i(0)\) randomly oriented, satisfying \(\|\mathbf{v}_i(0)\| \leq1\, \text{m/s}\). The interaction radius is \(r_i = 7.5\, \text{m}\), \(k = 0.1\), \( v_i^{\max} = 2.5\, \text{m/s}\), and \(u_i^{\max} = 5\; \mathrm{m/s^2}\). 

We measure directional alignment using the metric
\[
\gamma = \frac{1}{n}\sum_{i=1}^n \frac{1}{|\mathcal{N}_i|}\sum_{j \in \mathcal{N}_i}\frac{\mathbf{v}_i^\top \mathbf{v}_j}{\|\mathbf{v}_i\|\|\mathbf{v}_j\|},
\]
where the inner term computes the average cosine similarity between agent \( i \)’s velocity and its neighbors’ velocities, and the outer term averages across all agents. The metric \( \gamma \in [-1, 1] \), with \( \gamma \approx 1 \) indicating near-parallel velocities (strong alignment) and values near or below 0 indicating misalignment. By normalizing direction, this metric isolates directional consensus from speed differences.

Fig.~\ref{fig:trajectory} shows the collective trajectory under the velocity-alignment-based model \eqref{eq:control-flk} (V-based), the position-based model \eqref{eq:position-model} with the threshold alignment gain (P-based (thr.)) and no threshold alignment gain (P-based (no thr.)) over \([0,100] \, \text{s}\). 

For the velocity-alignment-based (V-based) model, the trajectories show that, starting from random initial conditions, the flock achieves cohesive and coherent motion with relatively strong alignment. However, the collective direction continues to vary over time, resulting in a spatially flexible flocking pattern without persistent heading. In contrast, the position-based model with threshold (P-based (thr.)) produces trajectories that are both cohesive and coherent and exhibit a clearly persistent collective direction throughout the simulation, forming a more compact and directionally stable flocking configuration. 

To illustrate the critical role of the threshold gain $  \phi = k \, |\mathcal{N}_i|  $, we also simulate the position-based model without the threshold (P-based (no thr.)). These trajectories initially display strong alignment and persistent collective direction similar to the threshold case; however, over an extended time, the alignment gradually weakens as the gain decays, leading to a noticeable loss of directional coherence while spatial cohesion is largely maintained.

Fig. \ref{fig:hist} tracks (a) the alignment metric $\gamma$, (b) inter-agent distances, (c) average flock speeds, and (d) the cohesive radius of the flock (defined as the maximum distance from any agent to the flock centroid). For the velocity-alignment-based (V-based) model, the alignment metric $  \gamma  $ reaches a peak around $  t \approx 30\,\text{s}  $ and subsequently exhibits slight oscillations. This is consistent with the spatially flexible collective motion and varying direction observed in the trajectory plots. In contrast, the position-based model with threshold (P-based (thr.)) exhibits rapid convergence to near-maximal alignment ($  \gamma \approx 1  $) within the transient phase, with this strong alignment sustained throughout the simulation. It also maintains smaller and more stable inter-agent separations (typically \(1.5-2\,\text{m}\)) compared to the V-based model (\(2-3\,\text{m}\)), resulting in a visibly more compact formation with minimal variation in distances over time. For the position-based model without threshold (P-based (no thr.)), $  \gamma  $ shows rapid initial convergence to strong alignment but gradually declines as time progresses due to the decaying gain. This leads to a progressive loss of directional coherence over extended periods, confirming that the non-vanishing threshold is essential for maintaining persistent collective alignment and flock identity.

\begin{figure}
    \centering
    \includegraphics[width=1\linewidth]{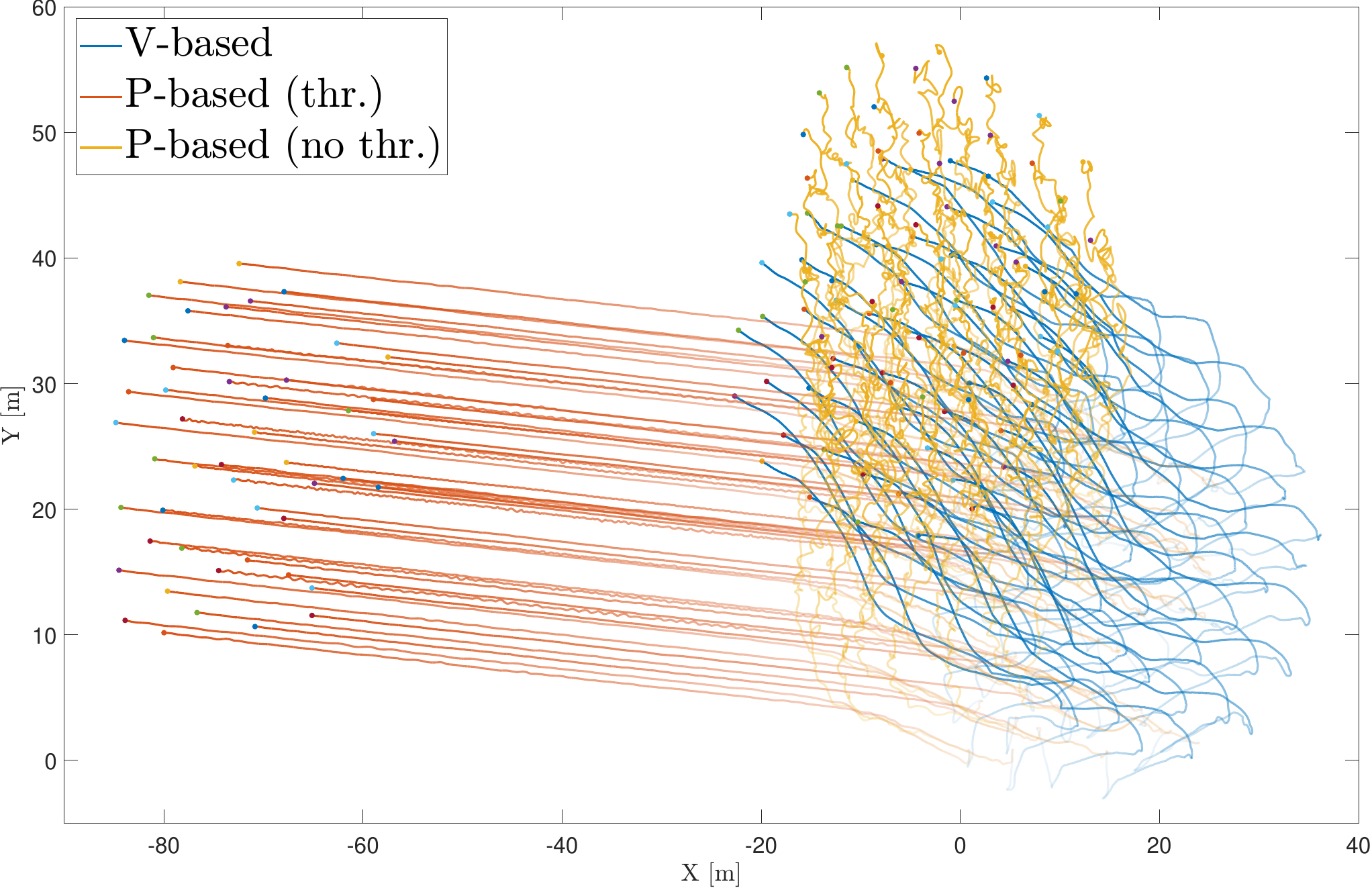}
    
    \caption{Trajectories of the collective over \([0, 100]\,\text{s}\) under (a) the velocity-alignment-based model \eqref{eq:control-flk} (V-based), (b) the position-based model \eqref{eq:position-model} with threshold alignment gain (P-based (thr.)), and (c) the position-based model without threshold (P-based (no thr.)), all starting from identical random initial conditions. The V-based trajectories achieve cohesive and coherent motion with relatively strong alignment, yet exhibit a spatially flexible pattern with varying collective direction over time. The P-based (thr.) trajectories rapidly converge to strong, persistent alignment and form a compact, directionally stable flocking configuration throughout the simulation. The P-based (no thr.) trajectories initially display strong alignment and persistent direction similar to the threshold case but show gradual weakening of alignment over extended time due to the decaying gain, resulting in progressive loss of directional coherence.}
    \label{fig:trajectory}
\end{figure}

\begin{figure}
    \centering
    \begin{subfigure}{0.45\textwidth}
        \centering
        \includegraphics[width=\linewidth]{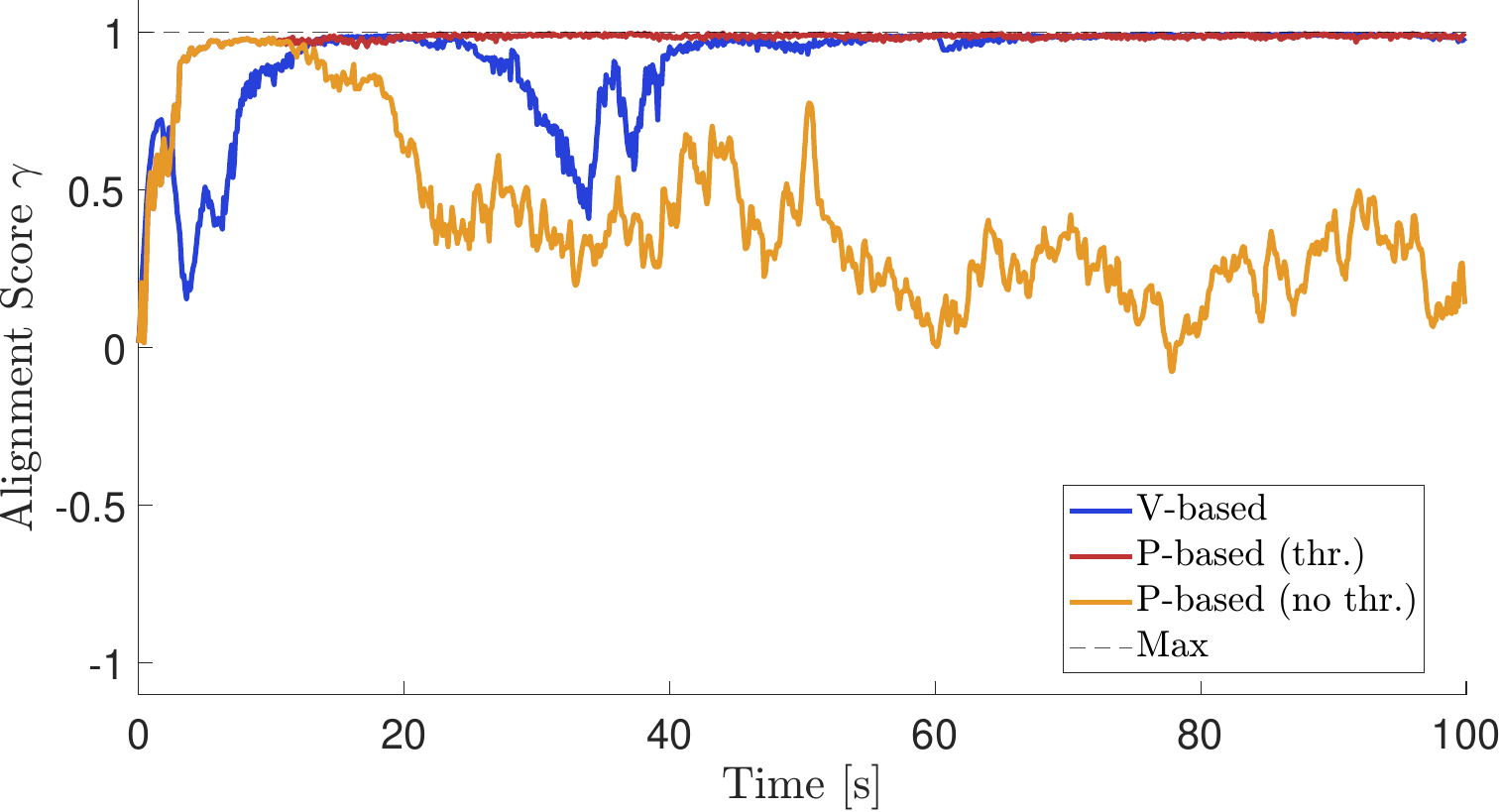}
        \caption{}
    \end{subfigure}
    \begin{subfigure}{0.45\textwidth}
        \centering
        \includegraphics[width=\linewidth]{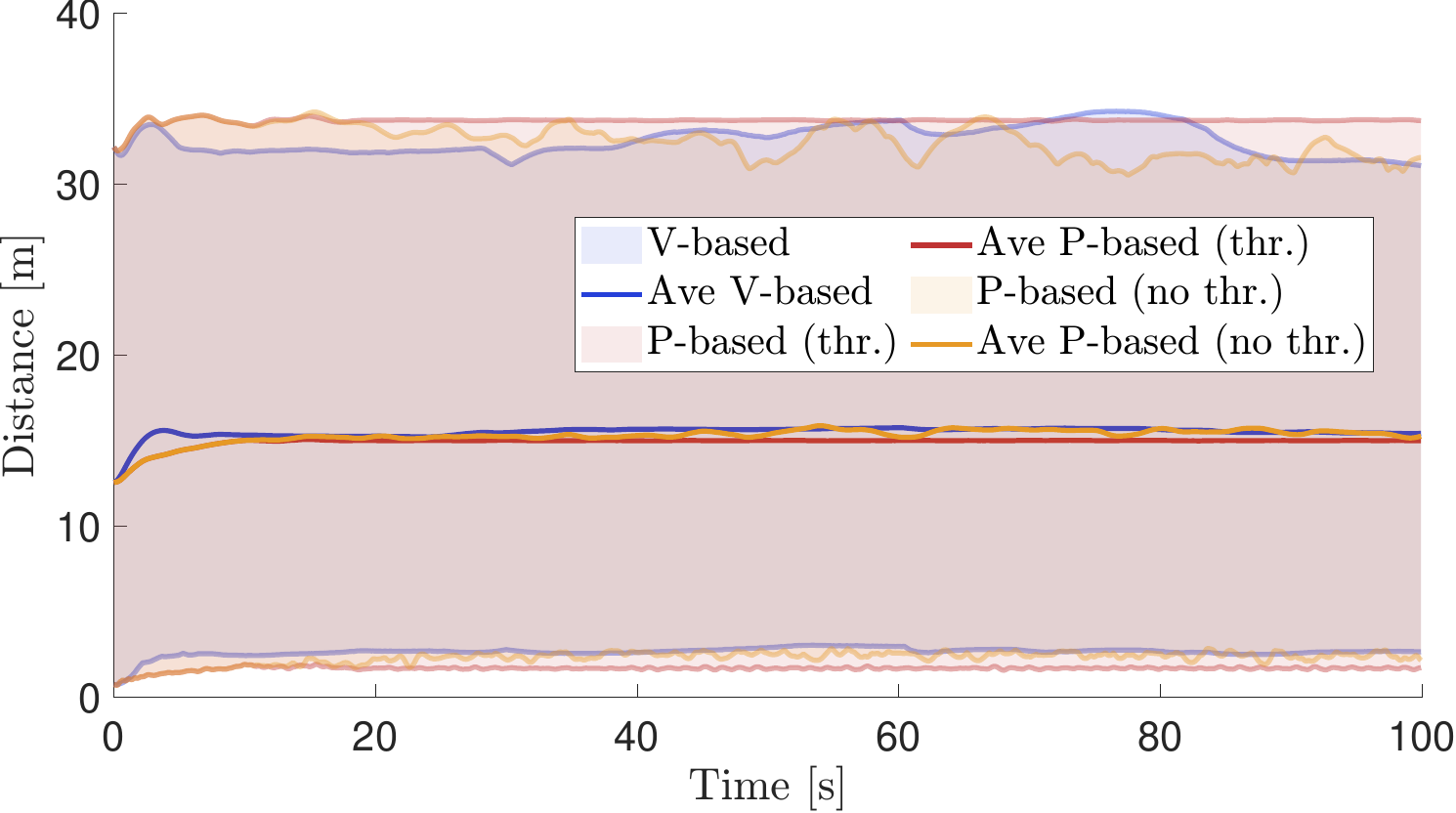}
        \caption{}
    \end{subfigure}
    \begin{subfigure}{0.45\textwidth}
        \centering
        \includegraphics[width=\linewidth]{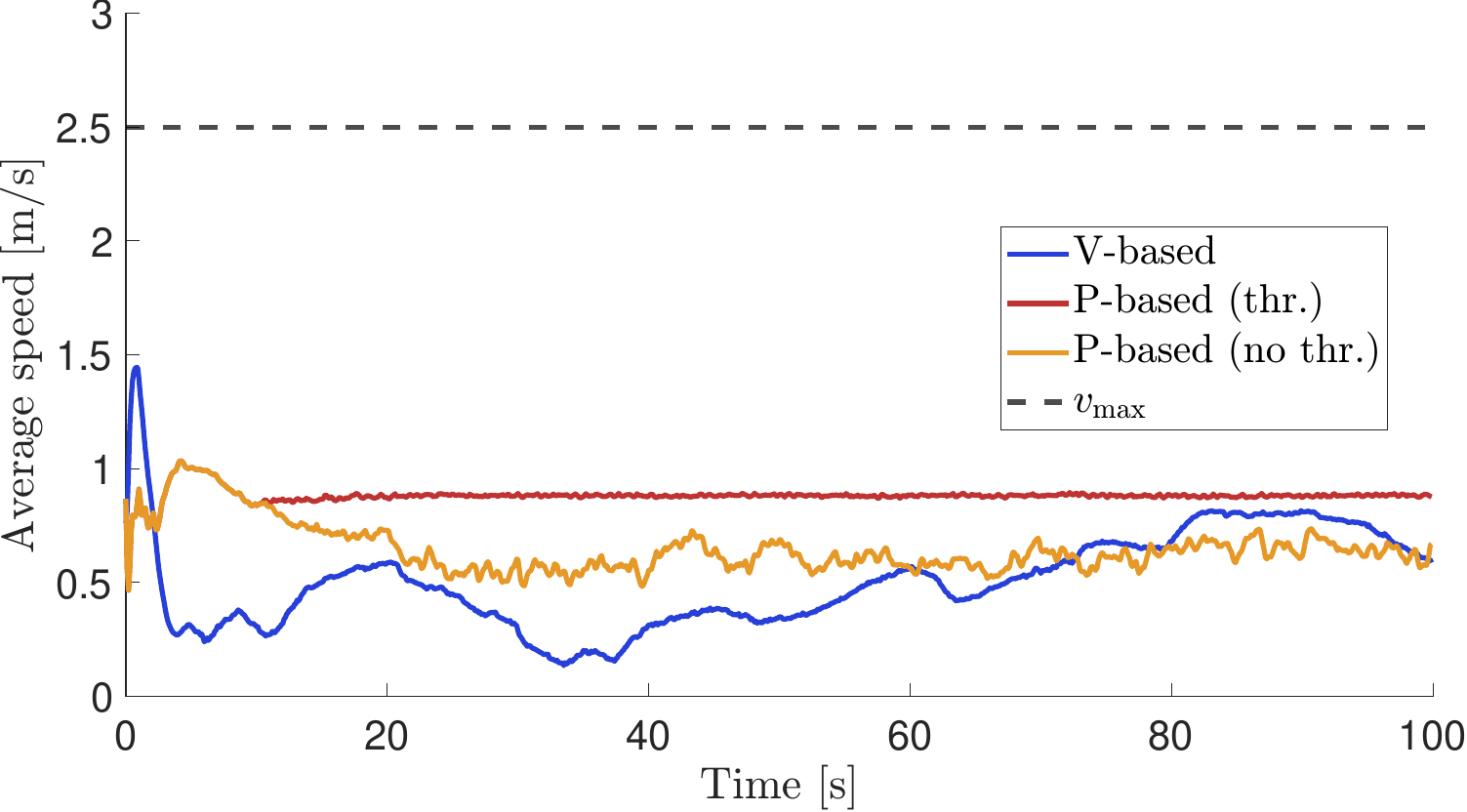}
        \caption{}
    \end{subfigure}
    \begin{subfigure}{0.45\textwidth}
        \centering
        \includegraphics[width=\linewidth]{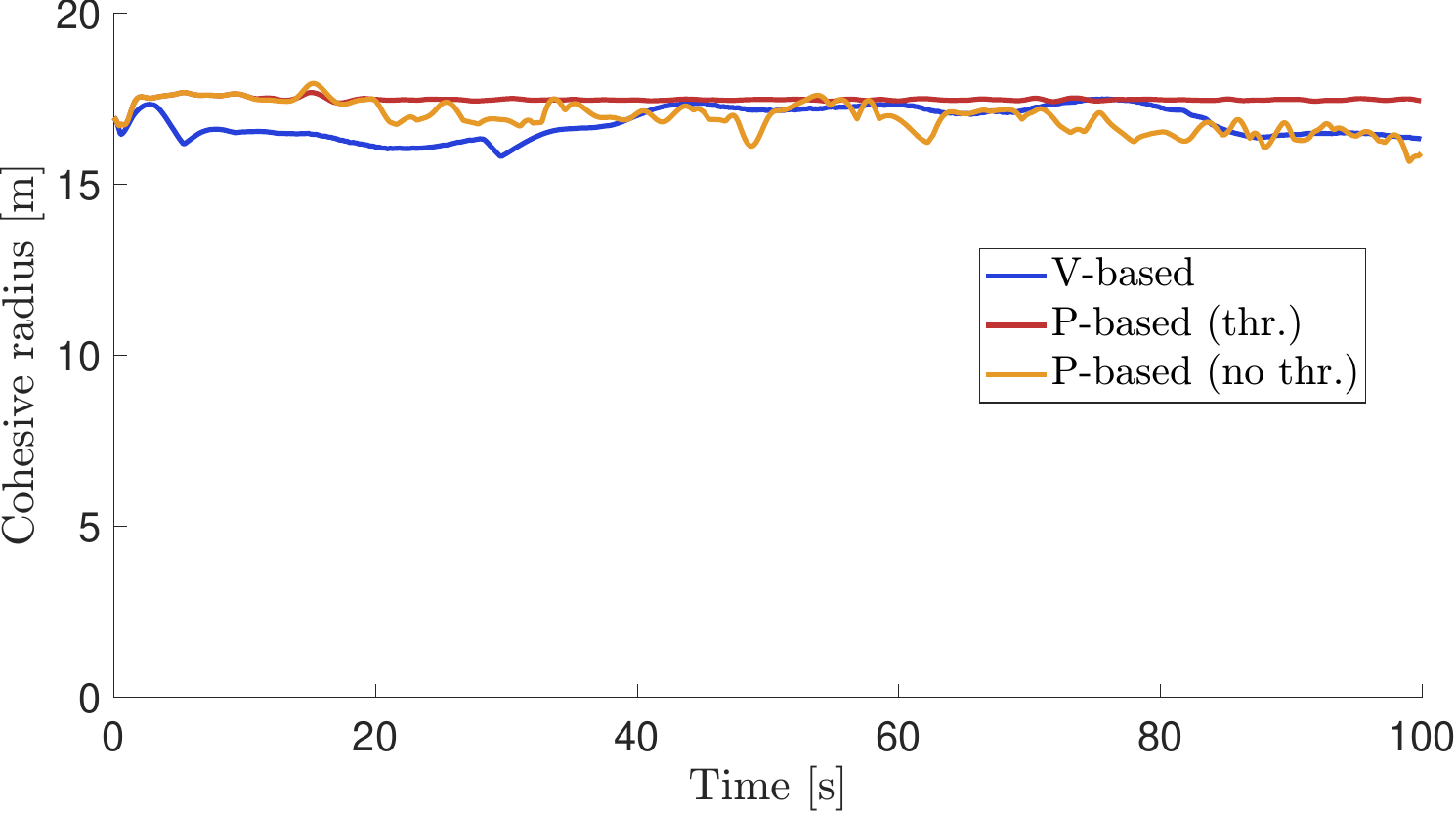}
        \caption{}
    \end{subfigure}
    \caption{Time histories for the collective trajectory under the velocity-alignment-based model \eqref{eq:control-flk} (V-based), the position-based model \eqref{eq:position-model} with the threshold alignment gain (P-based (thr.)) and no threshold alignment gain (P-based (no thr.)): (a) alignment metric \( \gamma \), (b) inter-agent distances, (c) average speeds, and (d) collective radii. }
    \label{fig:hist}
\end{figure}

Table~\ref{tab:flocking-comparison2} provides a qualitative comparison between the velocity-alignment-based model (V-based) and the proposed position-based model with threshold gain (P-based (thr.)). The position-based approach achieves markedly stronger and more persistent alignment, along with narrower and more stable inter-agent separations, resulting in superior cohesion and directional consensus. However, this comes at the cost of a more compact formation structure with limited spatial flexibility. In contrast, the V-based model produces a spatially adaptable, flexible configuration with wider separations and slightly oscillating but still relatively strong alignment. Biologically, the compact, directionally stable formations produced by the position-based model closely resemble the highly ordered, persistent V- or J-shaped structures observed in migratory bird flocks during long-distance travel. Conversely, the more flexible and spatially adaptive configurations of the V-based model are consistent with the dynamic, loosely coordinated patterns seen in foraging groups or during predator-avoidance maneuvers in fish schools and bird flocks.

\begin{table*}
\caption{Qualitative comparison of key flocking characteristics between the velocity-alignment-based model (V-based) and the proposed position-based model with threshold (P-based (thr.)).}
\label{tab:flocking-comparison2}
\centering
\begin{tabular}{|l|c|c|}
\hline
\textbf{Characteristic}          & \textbf{V-based}              & \textbf{P-based (thr.)}       \\
\hline
Alignment strength \& persistence & Strong, with minor variation   & Very strong \& persistent     \\
\hline
Formation structure              & Flexible, spatially adaptive  & Less flexible, directionally stable   \\
\hline
Inter-agent separation           & Wider      & Narrower \\
\hline
Cohesion \& directional consensus & Moderate                     & Superior                      \\
\hline
\end{tabular}
\end{table*}

To assess whether the flock preserves a rigid initial formation, we compute the variance of pairwise distances $\mathrm{Var}(\|\mathbf{p}_j - \mathbf{p}_i\|)$, which serves as a direct quantitative measure of formation rigidity. As shown in Fig.~\ref{fig:metrics}, this variance increases during the transient phase, indicating that exact preservation of initial inter-agent distances is not enforced. Nevertheless, the cohesion radius remains nearly constant, confirming sustained group cohesion without rigid adherence to the initial formation. This behavior suggests that the initial relative positions function as a short-term memory imprint that drives rapid early alignment and cohesion but gradually fades once the alignment gain stabilizes. The position-based model thereby exploits this transient memory mechanism to achieve fast convergence to strong and persistent collective alignment.
\begin{figure}
\centering
\includegraphics[width=0.5\linewidth]{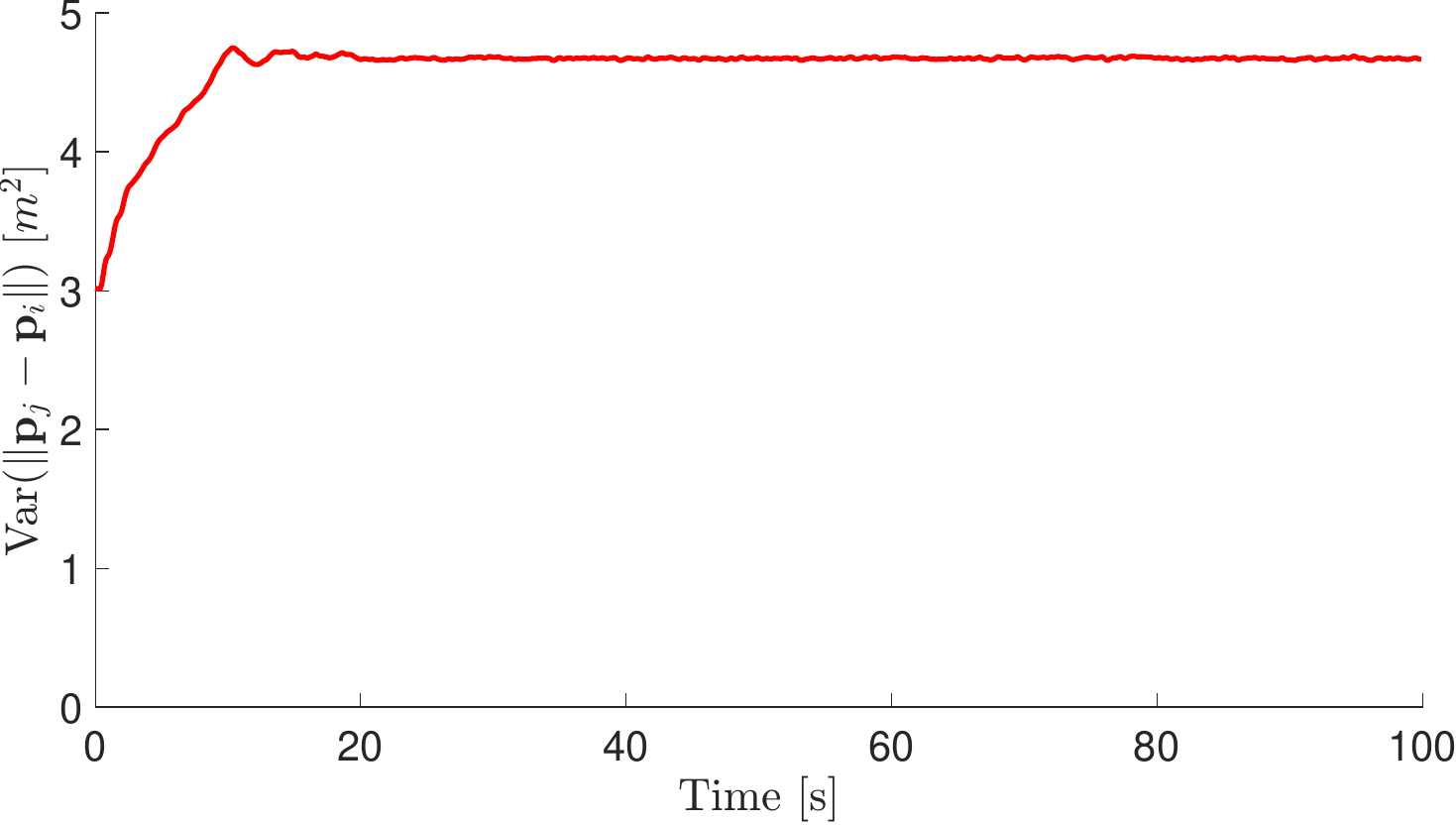}
\caption{Assessment of formation rigidity. Pairwise distance variance increases during the transient phase, showing that the initial formation is not preserved.}
\label{fig:metrics}
\end{figure}

\section{Experiment}\label{sec:experiment}

The position-based flocking model was evaluated using nine GRITSBot wheeled robots (dimensions: \(11\times 10\times7\,\text{cm}\)) in the Robotarium testbed~\cite{wilson2020robotarium}. Robots were initialized at random positions and headings within a \(1.0 \times 1.0\,\text{m}\) region inside the \(3.2 \times 2.0\,\text{m}\) arena. Each robot interacted within a radius \(r_i = 0.75\,\text{m}\) according to \eqref{eq:position-model}, with \(\delta_i = 0.12\), \(k = 0.15\), \(v_i^{\max} = 0.15\,\text{m/s}\), \(u_i^{\max} = 0.5\,\text{m/s}^2\). Robot positions were tracked using the overhead Vicon motion capture system, and control inputs were computed every \(0.033\,\text{s}\) over a \(120 \,\text{s}\) experiment. Double-integrator dynamics from \eqref{eq:position-model} were mapped to unicycle commands, with target linear and angular velocities conservatively saturated at \(0.15\,\text{m/s}\) and \(0.55\,\text{rad/s}\), respectively, to accommodate the compact arena.

Snapshot sequences in Fig.~\ref{fig:trajectory_snapshots} illustrate the emergence of cohesive and coherent flocking. The alignment metric in Fig.~\ref{fig:ali_exp} confirms the establishment and maintenance of collective motion direction. Actuation inaccuracies of the Robotarium platform introduced minor fluctuations; however, coherent flocking behavior was consistently observed.

\begin{figure*}
    \centering
    \includegraphics[width=0.32\linewidth]{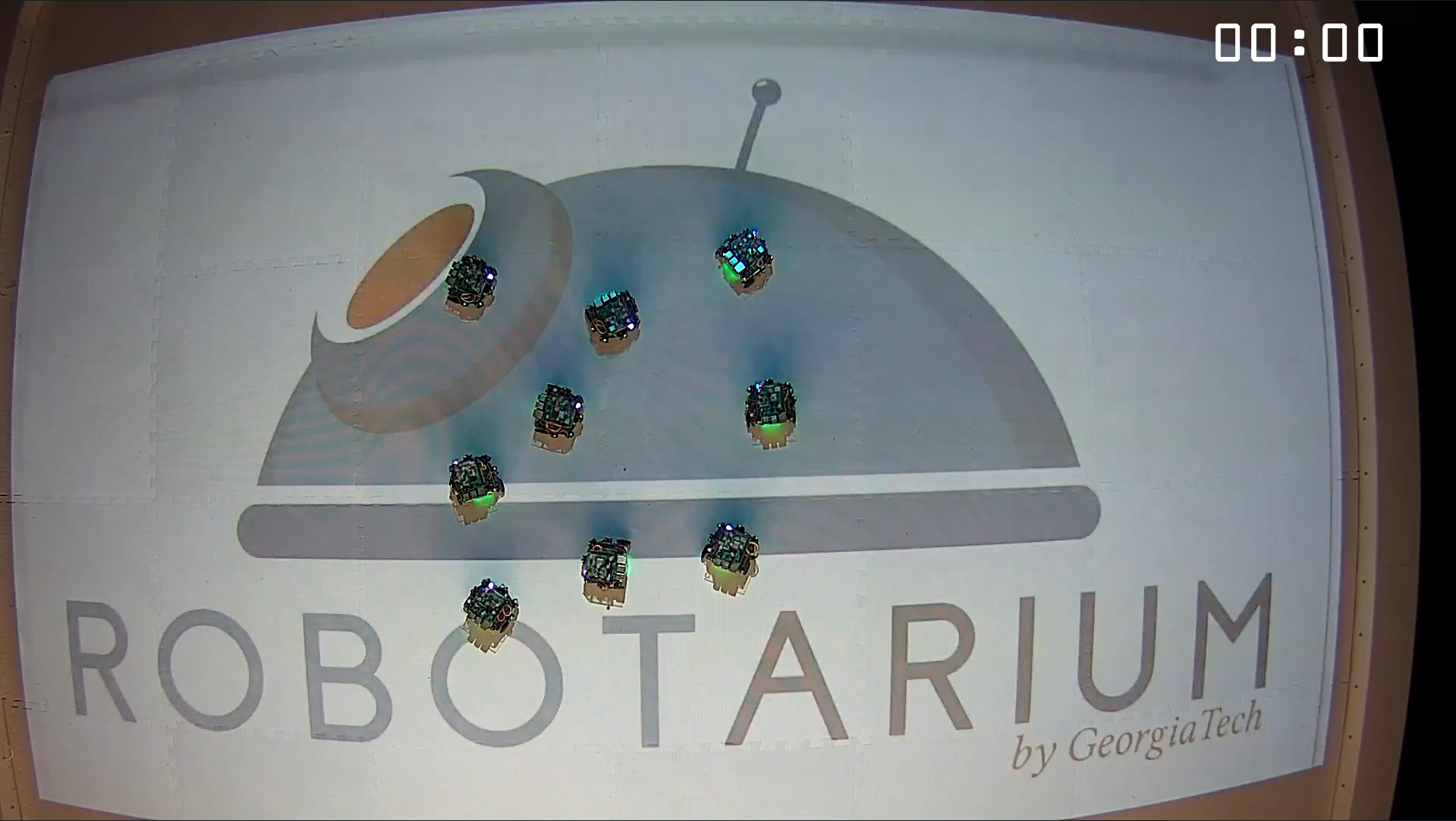}
    \includegraphics[width=0.32\linewidth]{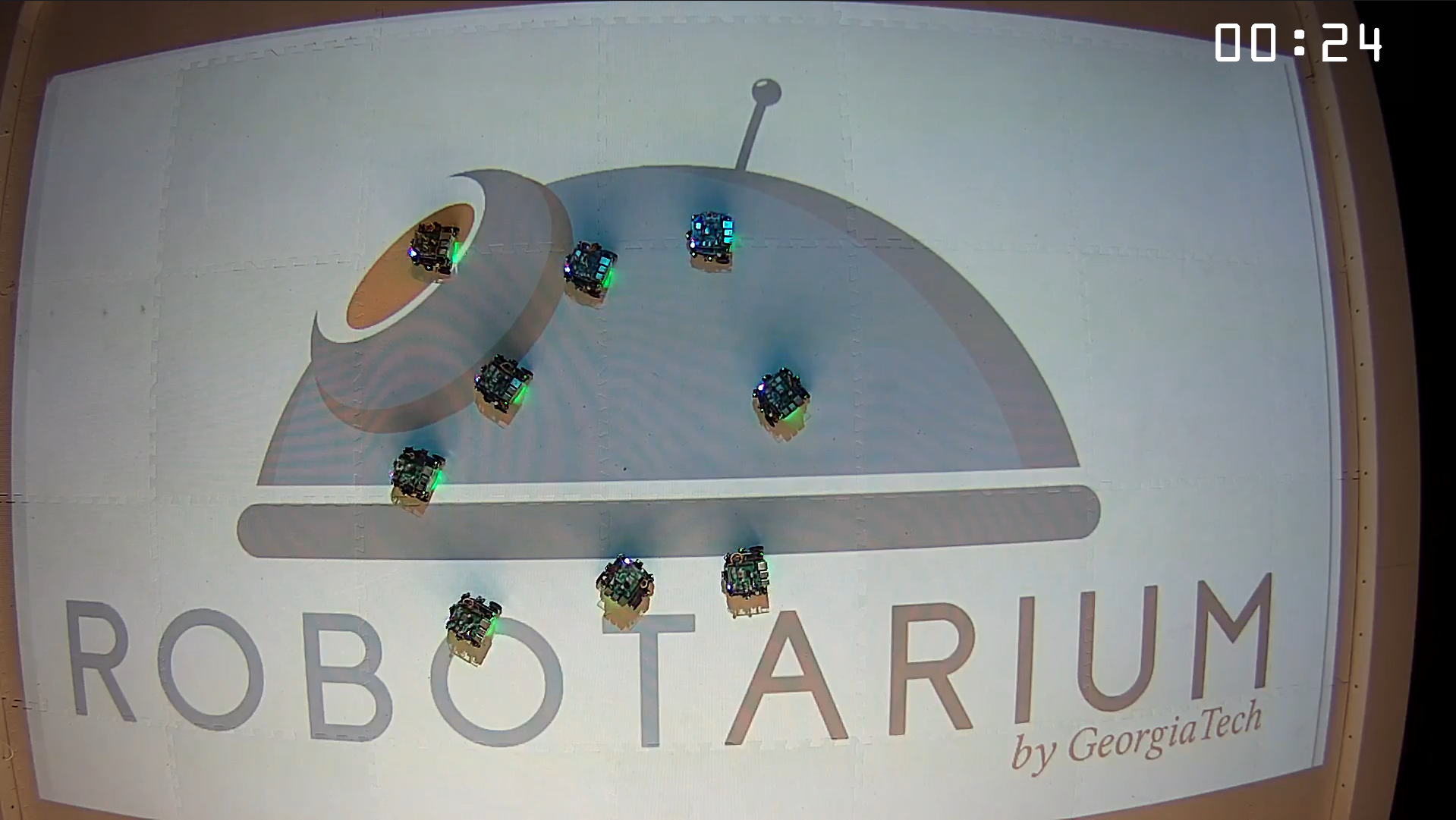}
    \includegraphics[width=0.32\linewidth]{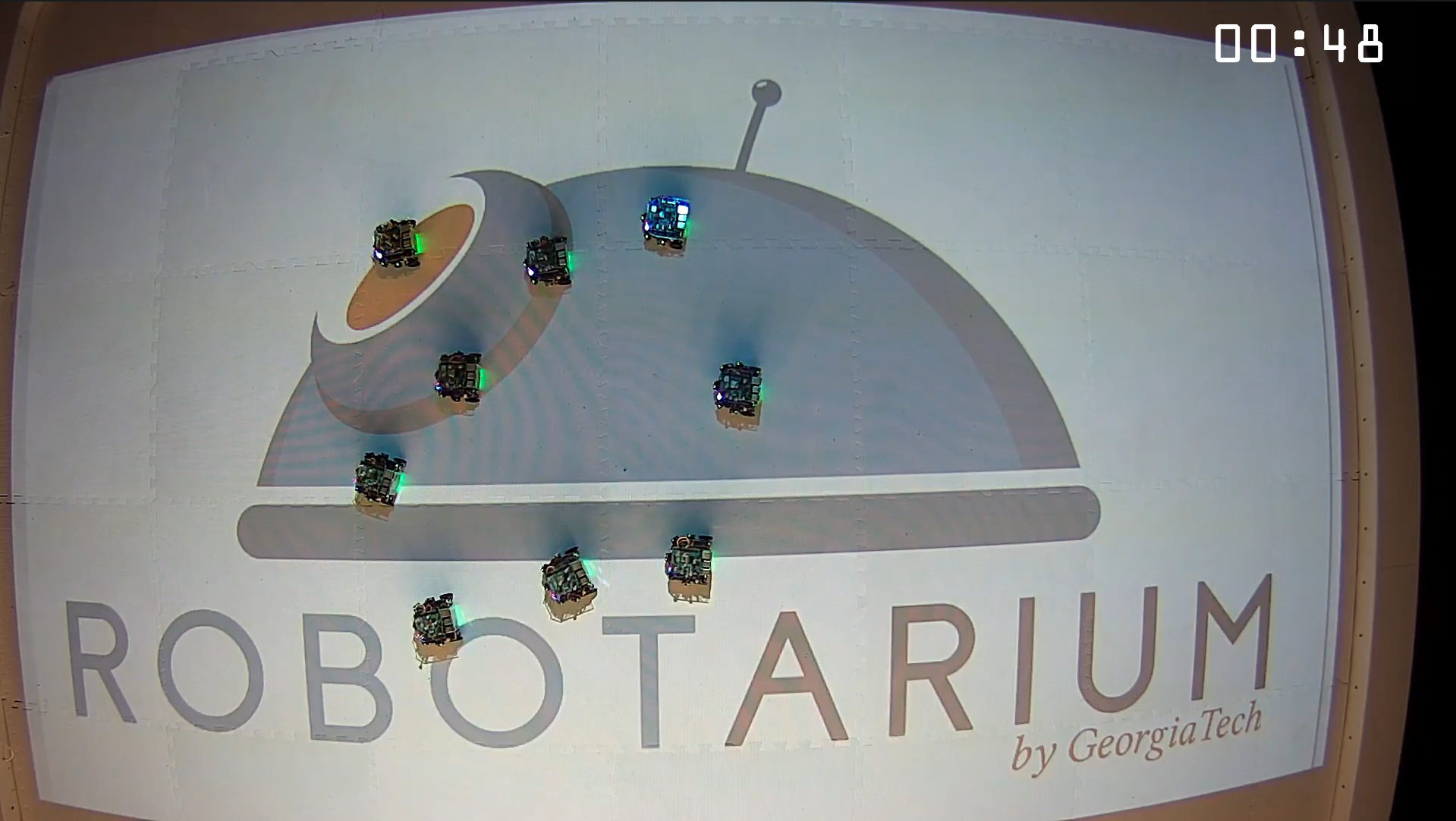}
    \includegraphics[width=0.32\linewidth]{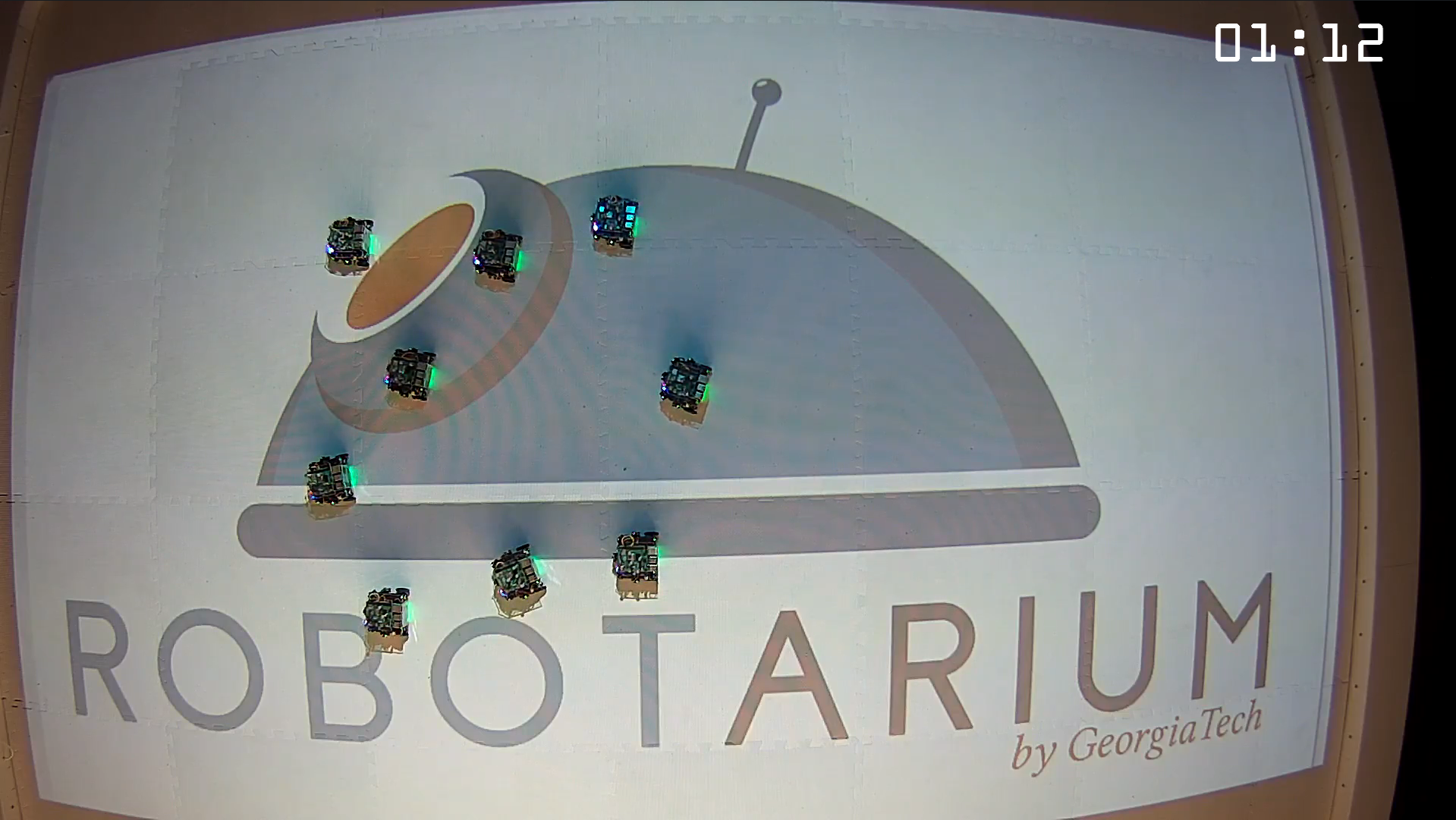}
    \includegraphics[width=0.32\linewidth]{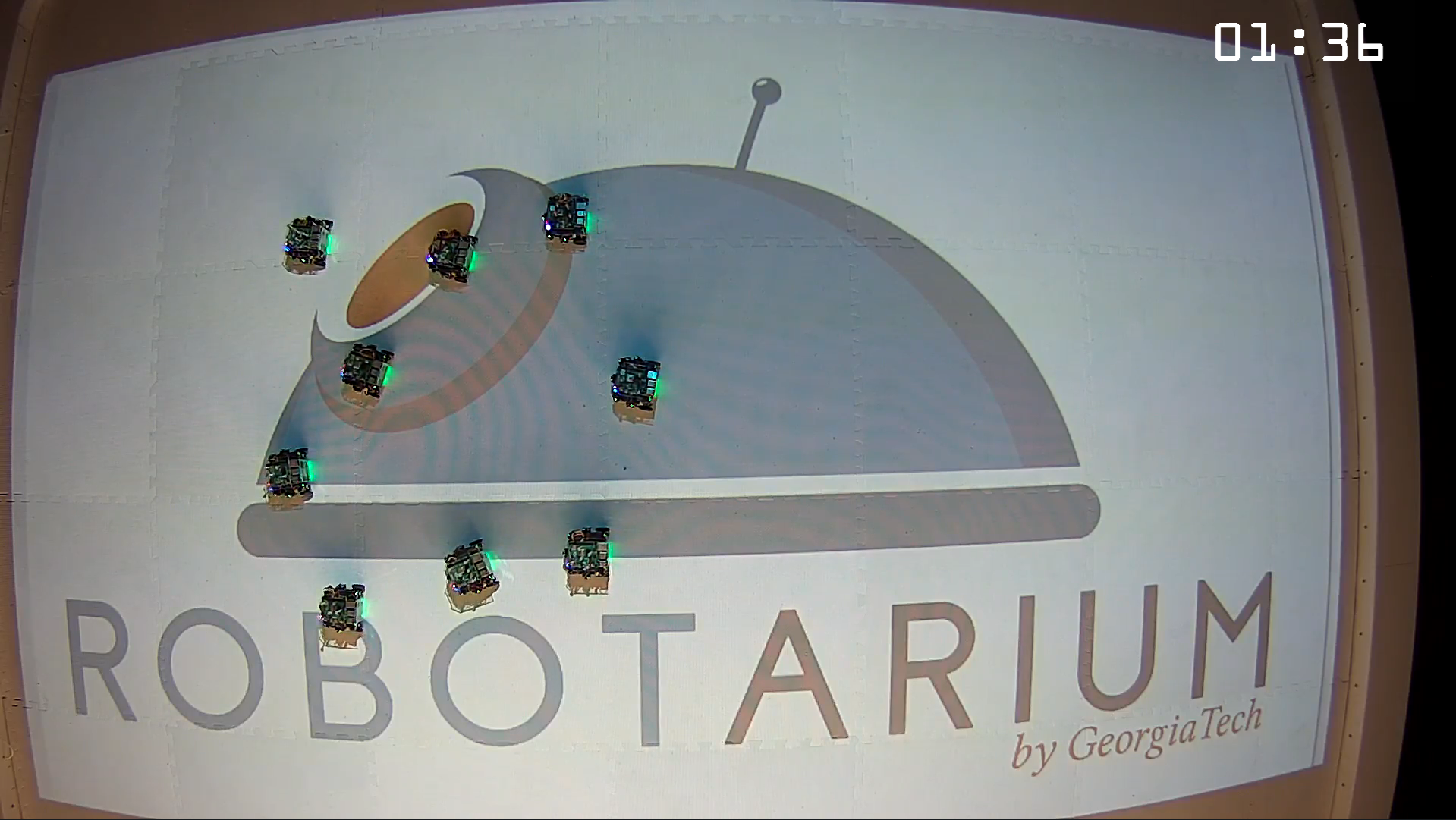}
    \includegraphics[width=0.32\linewidth]{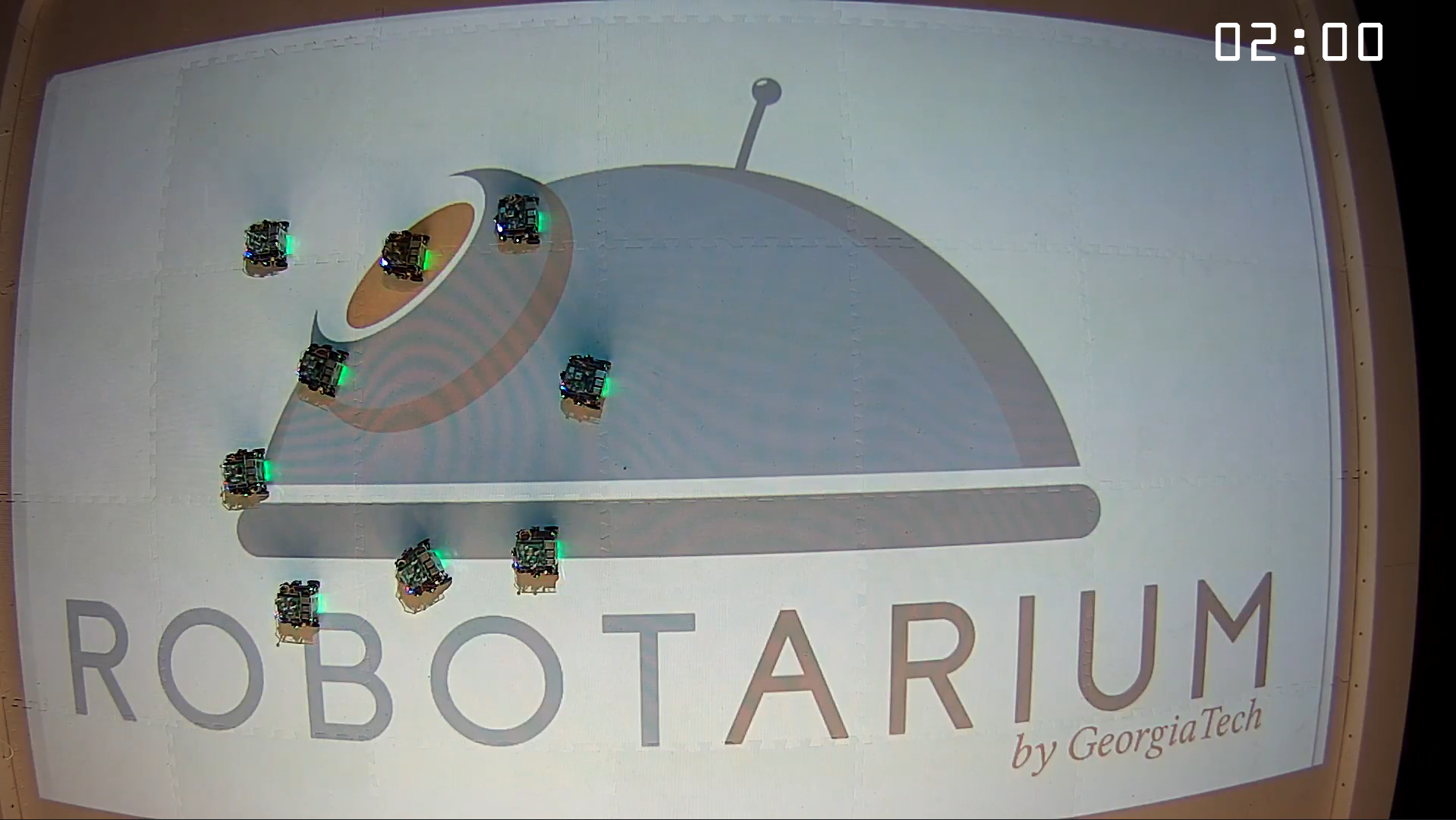}
    \caption{Snapshots from the real-robot experiment demonstrating cohesive and coherent flocking under the position-based model. A supplementary video showing the full experiment, including real-time data visualization and trajectory evolution, is available at \url{https://youtu.be/g8ClSYKpmQY}.}
    \label{fig:trajectory_snapshots}
\end{figure*}

\begin{figure}
    \centering
    \includegraphics[width=0.5\linewidth]{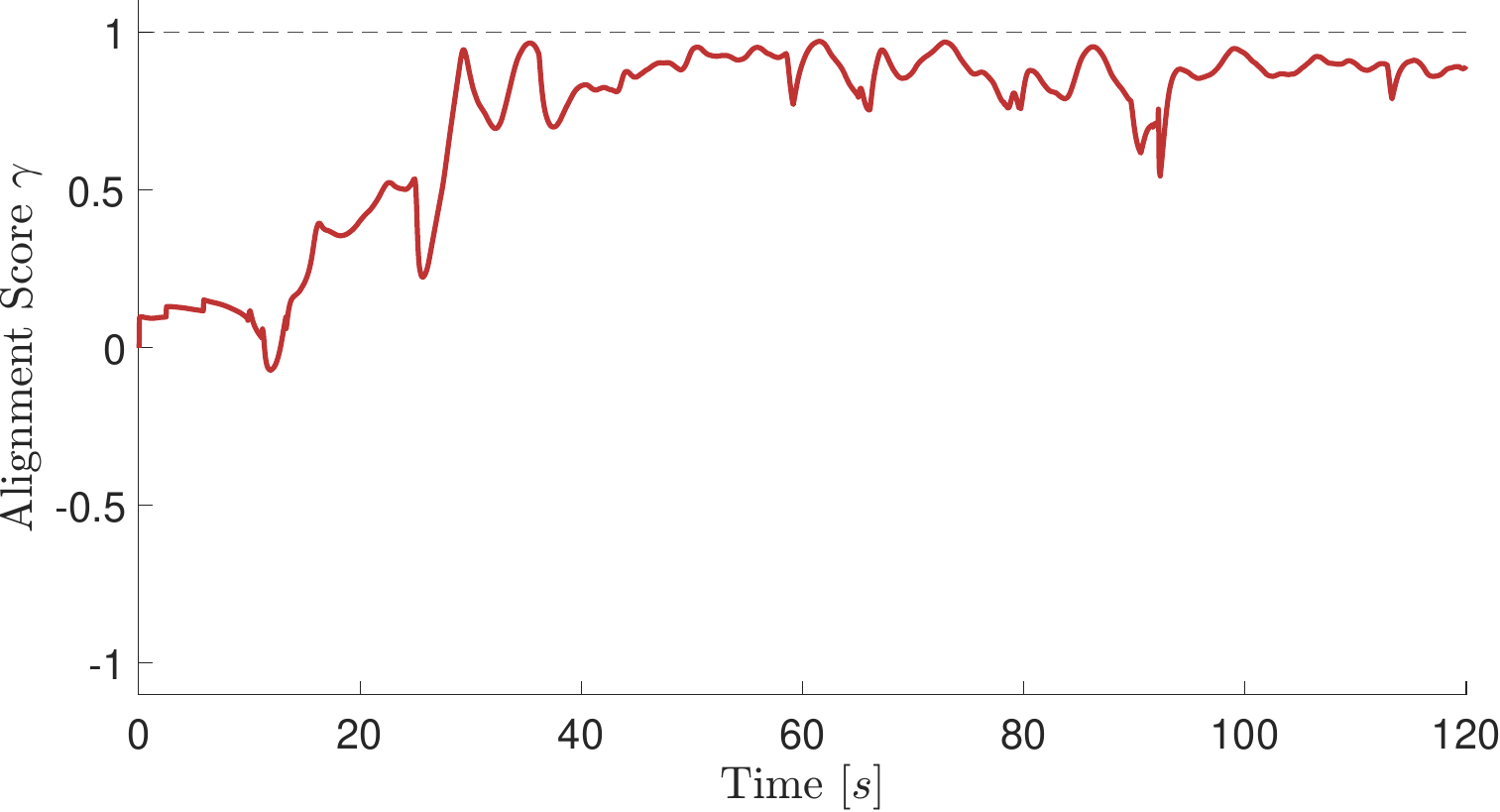}
    \caption{Alignment metric in the experiment.}
    \label{fig:ali_exp}
\end{figure}

\section{Conclusion}\label{sec:conclusion}

This paper presents a position-based flocking model that achieves persistent velocity alignment using only relative positions. By approximating velocity differences from position changes and applying a non-vanishing threshold gain, the model prevents alignment decay. Simulations and real-robot experiments show stronger, more persistent alignment and more compact formations with directionally stable structures emerging from local interactions alone. Biologically, these stable formations with persistent collective direction resemble migratory bird flocks. The approach offers a robust, position-only solution for sensing- and communication-constrained robotic swarms. Future work will address mechanisms for introducing spatial flexibility in confined and heterogeneous settings, better matching biological foraging and evasive flocking behaviors.

\subsection*{Acknowledgments}
This work was partially funded by the Czech Science Foundation (GAČR) under research project no. 23-07517S and the European Union under the project Robotics and Advanced Industrial Production (reg. no. CZ.02.01.01\slash00\slash22\_008\slash0004590). This work was also supported in part by The Scientific and Technological Research Council of Türkiye (TÜBİTAK).

\bibliographystyle{unsrt}
\bibliography{mybibliography}

\end{document}